\documentclass[10pt,twocolumn,letterpaper]{article}

\usepackage{iccv}
\usepackage{times}
\usepackage{epsfig}
\usepackage{graphicx}
\usepackage{amsmath}
\usepackage{amssymb}
\usepackage{cite}
\usepackage{mathrsfs}
\usepackage{color}
\usepackage[accsupp]{axessibility}

\usepackage[breaklinks=true,bookmarks=false]{hyperref}

\iccvfinalcopy 

\def\iccvPaperID{****} 

\ificcvfinal\fi

\begin{document}

\title{Memory Based Video Scene Parsing}

\author{Zhenchao Jin$^{1}$, Dongdong Yu$^{2}$, Kai Su$^{2}$, Zehuan Yuan$^{2}$, Changhu Wang$^{2}$\\
$^{1}$University of Science and Technology of China \quad 
$^{2}$ByteDance \\
{\tt\small blwx@mail.ustc.edu.cn} \\
{\tt\small \{yudongdong, sukai, yuanzehuan, wangchanghu\}@bytedance.com},
}

\maketitle
\ificcvfinal\thispagestyle{empty}\fi

\begin{abstract}
   Video scene parsing is a long-standing challenging task in computer vision, aiming to assign pre-defined semantic labels to pixels of all frames in a given video.
   Compared with image semantic segmentation, this task pays more attention on studying how to adopt the temporal information to obtain higher predictive accuracy.
   In this report, we introduce our solution for the 1st Video Scene Parsing in the Wild Challenge, which achieves a mIoU of 57.44 and obtained the 2nd place (our team name is CharlesBLWX).
\end{abstract}

\section{Overview}
Our goal is to build an accurate 2D segmentor for video scene parsing.
As the performance is the most important term for this challenge, we adopt BEiT \cite{bao2021beit} as our backbone network, \emph{i.e.}, the encoder structure.
Then, we take the upernet decoder \cite{xiao2018unified} as our baseline method and introduce several improvements to build the final video semantic segmentation framework.
Below we first present the main idea of our method in Section \ref{approach},
followed by the network structure used in the proposed segmentor in Section \ref{structure}.
Then, we describe the train and inference strategy used to further boost our segmentation performance in Section \ref{strategy}.
Finally, we report the experiments and detailed ablation analysis in Section \ref{exp}.

\section{Our Approach} \label{approach}
Although video semantic segmentation is different from image semantic segmentation, 
a stronger image semantic segmentor usually can achieve the better performance in video semantic segmentation.
Therefore, we adopt BEiT backbone network \cite{bao2021beit} and upernet decoder \cite{xiao2018unified} as our baseline method, whose performance on ADE20K \cite{zhou2017scene} has achieved a state-of-the-art mIoU, \emph{i.e.}, $57.00\%$.

Based on this, we improve the decoder structure by using two kinds of memory strategies.
One is to set up a feature memory module and store the dataset-level representations of various classes in it. Then, we adopt these dataset-level representations to augment the pixel representations of current input image.
The other is to incorporate the temporal memory attention module (\emph{i.e.}, TMA module) \cite{wang2021temporal} into the first decoder structure.

After training both segmentors on VSPW \cite{miao2021vspw}, we ensemble the output class probability distributions of the segmentors to obtain the final prediction results.

\section{Model Structure} \label{structure}
\noindent \textbf{Encoder Structure.} We leverage BEiT \cite{bao2021beit} as our encoder structure, in which we set $layer=24$, $hidden=1024$, $FFN factor=4\times$, $head=16$ and $patch=16 \times 16$.
The initialized weights are self-supervised pretrained and then intermediate fine-tuned on ImageNet22k following \cite{bao2021beit}.

\begin{figure*}
\centering
\includegraphics[width=0.95\textwidth]{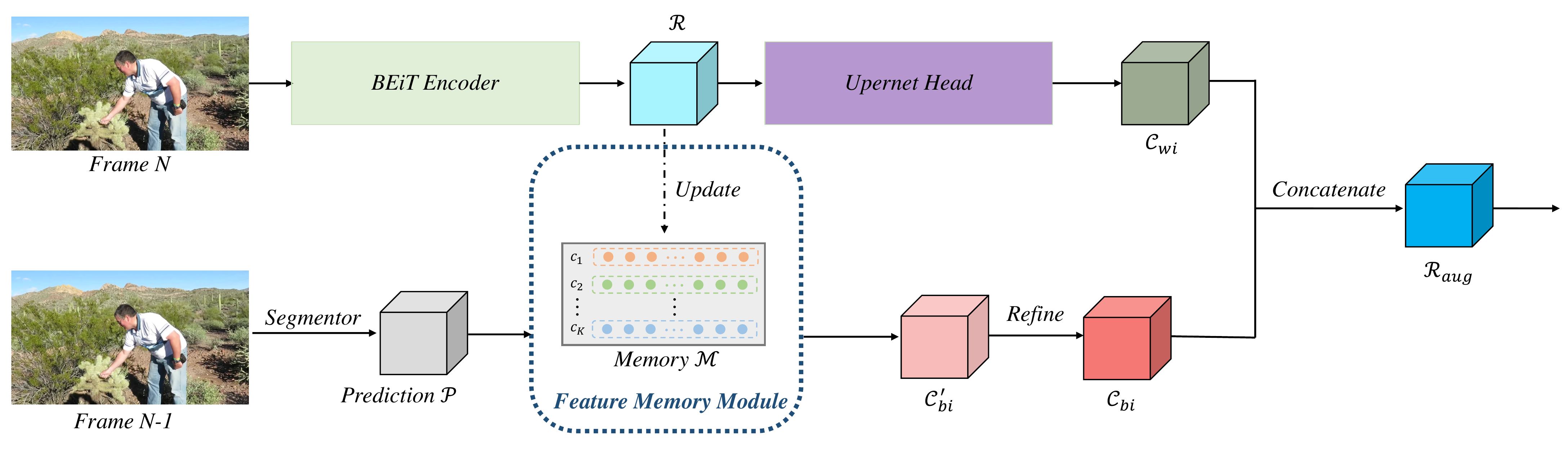}
\caption{
   Illustrating the pipeline of Decoder A. 
}\label{framework}
\end{figure*}

\begin{figure*}
\centering
\includegraphics[width=0.95\textwidth]{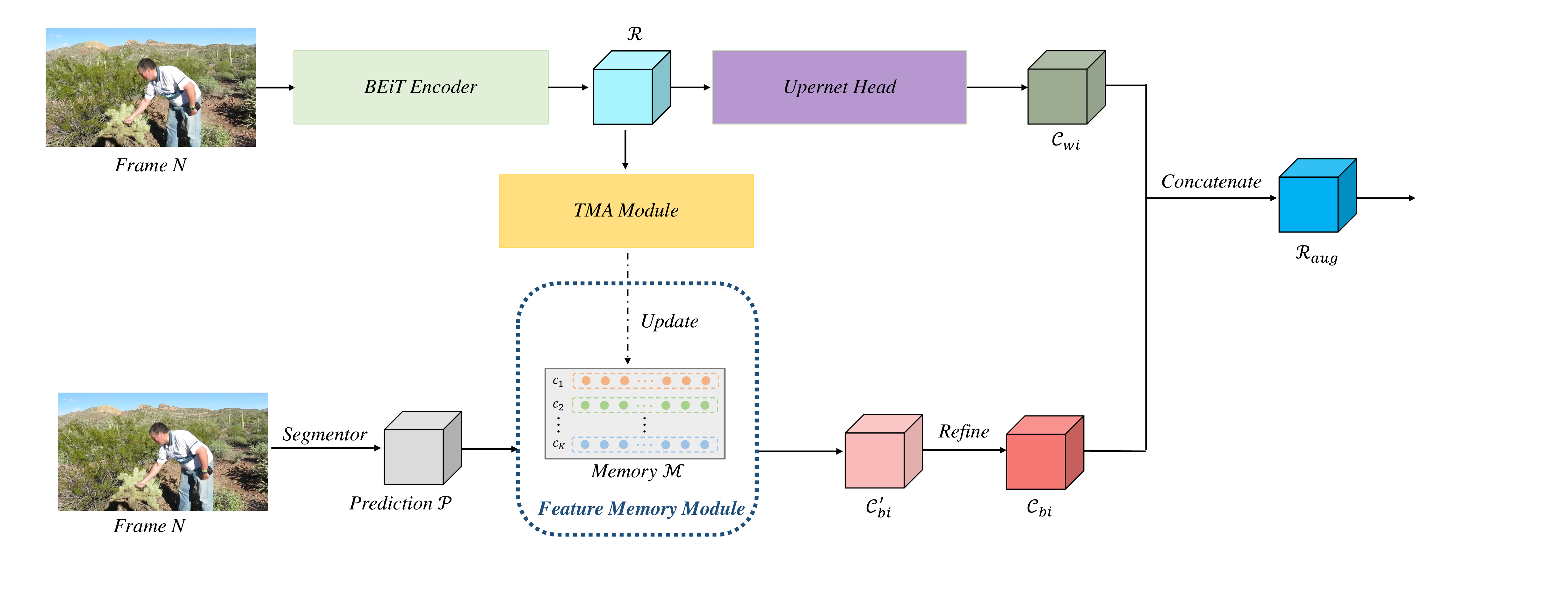}
\caption{
   Illustrating the pipeline of Decoder B. 
}\label{framework2}
\end{figure*}

\noindent \textbf{Decoder Structure A.} 
Here we introduce the first decoder structure designed in this paper.
This structure is based on our paper \emph{Mining Contextual Information Beyond Image for Semantic Segmentation} accepted by ICCV 2021 \cite{jin2021mining}.
As illustrated in Figure \ref{framework}, given the input $Frame~N$, we first leverage the BEiT backbone network to obtain the multi-level representations $\mathcal{R} = \{\mathcal{X}_1, \mathcal{X}_2, \mathcal{X}_3, \mathcal{X}_4\}$.
Then, we leverage the upernet head consisted of pyramid pooling module \cite{zhao2017pyramid} and feature pyramid network \cite{lin2017feature} to 
extract the basic representations $\mathcal{C}_{wi}$.
Next, we will set up a feature memory module $\mathcal{M}$ which is dynamically updated during training.
To be more specific, $\mathcal{M}$ of size $K \times C$ is introduced to store the dataset-level representations of various classes,
where $K$ is the number of the classes and the dimension of a pixel representation is $C$.
To set up the feature memory, we first randomly select one pixel representation for each category from the train set to initialize $\mathcal{M}$, where the representations are calculated by leveraging the backbone network BEiT.
Then, the values in $\mathcal{M}$ are updated by leveraging moving average after each training iteration:
\begin{equation} \label{eq1}
   \mathcal{M}_{t} = (1 - m_{t-1}) \cdot \mathcal{M}_{t-1} + m_{t-1} \cdot \mathscr{T}(\mathcal{R}_{t-1}),
\end{equation}
where $m$ is the momentum, $t$ denotes for the current number of iterations and $\mathscr{T}$ is used to transform $\mathcal{R}$ to have the same size as $\mathcal{M}$.

To implement $\mathscr{T}$, we first setup a matrix $\mathcal{R}'$ of size $K \times C$ and initialize it by using the values in $\mathcal{M}$. 
For the convenience of presentation, we leverage the subscript $[i, j]$ or $[i, *]$ to index the element or elements of a matrix. 
$\mathcal{R}$ is upsampled and permuted as size $HW \times C$, \emph{i.e.}, $\mathcal{R}^{HW \times C}$.
Subsequently, for each category $c_k$ existing in the input image, we have:
\begin{equation} \label{eq2}
   \mathcal{R}_{c_k} = \{\mathcal{R}^{HW \times C}_{[i, *]} ~|~  (\mathcal{GT}_{[i]} = c_k) \land (1 \leq i \leq HW) \},
\end{equation}
where $\mathcal{GT}$ of size $HW$ stores the ground truth category labels of $\mathcal{R}^{HW \times C}$.
$\mathcal{R}_{c_k}$ of size $N_{c_k} \times C$ stores the representations of category $c_k$ of $\mathcal{R}^{HW \times C}$.
$N_{c_k}$ is the number of pixels labeled as $c_k$ in the input image. Next, we calculate the cosine similarity matrix $\mathcal{S}_{c_k}$ of size $N_{c_k}$ between $\mathcal{R}_{c_k}$ and $\mathcal{M}_{[c_k, *]}$:
\begin{equation} \label{eq3}
   \mathcal{S}_{c_k} = \frac{\mathcal{R}_{c_k} \cdot \mathcal{M}_{[c_k, *]}}{{\lVert \mathcal{R}_{c_k} \rVert}_2 \cdot {\lVert \mathcal{M}_{[c_k, *]} \rVert}_2}.
\end{equation}
Finally, the representation of $c_k$ in $\mathcal{R}'$ is updated as:
\begin{equation} \label{eq4}
   \mathcal{R}'_{[c_k, *]} = \sum^{N_{c_k}}_{i=1} \frac{1 - S_{c_k, [i]}}{\sum^{N_{c_k}}_{j=1} (1 - S_{c_k,[j]})} \cdot \mathcal{R}_{c_k, [i, *]}.
\end{equation}
The output of $\mathscr{T}$ is $\mathcal{R}'$ which has been updated by all the representations of various classes in $\mathcal{R}^{HW \times C}$.

Then, $\mathcal{M}$ can be used to augment the original representations $\mathcal{R}$ of current input image by leveraging $\mathcal{P}$.
In our implementations, $\mathcal{P}$ is the predicted segmentation mask of previous frame, \emph{i.e.}, $Frame~N-1$.
Specifically, we select the dataset-level representations for each pixel according to $\mathcal{P}$ so that we can obtain $\mathcal{C}'_{bi}$.
Then, we calculate the relations between $\mathcal{R}$ and $\mathcal{C}'_{bi}$ so that we can obtain a position confidence weight to further refine $\mathcal{C}'_{bi}$.
Specifically, we first calculate the relations $\mathcal{O}$ as follows:
\begin{equation} \label{eq5}
   \mathcal{O} = Softmax(\frac{g_q(permute(\mathcal{R})) \otimes g_k(\mathcal{C}^{'}_{bi})^T}{\sqrt{\frac{C}{2}}}),
\end{equation}
where $permute$ is adopted to let $\mathcal{R}$ have size of $\frac{HW}{64} \times C$.
Then, $\mathcal{C}^{'}_{bi}$ is refined as follows:
\begin{equation} \label{eq6}
   \mathcal{C}_{bi} = permute(g_{o}(\mathcal{O} \otimes g_v(\mathcal{C}_{bi}^{'}))),
\end{equation}
where $g_q$, $g_k$, $g_v$ and $g_o$ are introduced to adjust the dimension of each pixel representation, implemented by a $1 \times 1$ convolutional layer.
$permute$ is used to let the output have size of $C \times \frac{H}{8} \times \frac{W}{8}$.

Finally, we concatenate the output of upernet head $\mathcal{C}_{wi}$ and the refined representations $\mathcal{C}_{bi}$ to predict the segmentation mask of current frame.

\noindent \textbf{Decoder Structure B.} 
Here we introduce the second decoder structure designed in this paper.
This decoder is similar to decoder structure A and there are two main differences:
\begin{itemize}
   \item We introduce the TMA module \cite{wang2021temporal} to better model the temporal information.
   Specifically, the input of feature memory module is the output of TMA rather than $\mathcal{R}$.

   \item $\mathcal{P}$ is the predicted segmentation mask of current frame $N$ rather than $N-1$.
\end{itemize}
After calculating $\mathcal{C}'_{bi}$, we also leverage the output of TMA to refine it so that we can obtain $\mathcal{C}_{bi}$.
Following decoder structure A, we finally concatenate the output of upernet head $\mathcal{C}_{wi}$ and the refined representations $\mathcal{C}_{bi}$ to predict the segmentation mask of current frame.

\noindent \textbf{Model Ensemble.} We simply add the outputs (\emph{i.e.}, the class probability distribution of each pixel representation) of both decoders to generate the final segmentation mask.

\begin{table}[t]
\centering
\caption{
   Ablation study of model ensemble on the validation set of VSPW. All the models are trained on the train set and tested under single-scale.
}\label{table1}
\resizebox{.45\textwidth}{!}{
\begin{tabular}{c|c|c|c}
   \hline
   \hline
   Method                              &Backbone       &Epochs                  &mIoU             \\
   \hline
   Upernet                             &BEiT-Large     &240                    &60.02            \\
   \hline
   Decoder A (\emph{ours})             &BEiT-Large     &240                    &61.24            \\
   Decoder B (\emph{ours})             &BEiT-Large     &240                    &61.18            \\
   Decoder A+B (\emph{ours})           &BEiT-Large     &240                    &\textbf{62.12}   \\
   \hline
   \hline
\end{tabular}}
\end{table}

\section{Train and Inference Strategy} \label{strategy}
\noindent \textbf{Train Strategy.} 
Here, we introduce several tricks used in training the proposed decoders.
\begin{itemize}
   \item In the initial stage of training, considering the instability of the segmentor,
   we replace $\mathcal{P}$ with the ground truth label of the selected frames.
   
   \item We use the data augmentaton to further boost the performance of our segmentor.
   Specifically, we adopt random scaling, horizontal flipping and color jitter following the default setttings in MMSegmentation \cite{mmseg2020}.

   \item Synchronized batch normalization implemented by pytorch is enabled during training. 
   
   \item  AdamW is used as our optimizer.
   
\end{itemize}

\noindent \textbf{Test Strategy.} 
Here, we introduce several tricks adopted in testing the proposed decoders.
\begin{itemize}
   \item We use multi-scale and flipping testing technology to obtain the best segmentation performance where the selected scales are $[0.75, 1.0, 1.25, 1.5, 1.75]$.
   \item To further boost the segmentation performance, we also save the predicted segmentation masks in the first-stage testing. And then, we replace $\mathcal{P}$ with the saved segmentation masks to performance the second-stage predicting.
   Similarly, we can perform the third-stage, fourth-stage testing and so on.
\end{itemize}

\begin{table}[t]
\centering
\caption{
   Ablation study of multi-stage inference on the validation set of VSPW. All the models are trained on the train set and tested under single-scale.
}\label{table2}
\resizebox{.42\textwidth}{!}{
\begin{tabular}{c|c|c|c}
   \hline
   \hline
   Method                              &Backbone       &Inference Stage                  &mIoU             \\
   \hline
   Decoder A                           &BEiT-Large     &stage0                    &61.24            \\
   \hline
   Decoder A                           &BEiT-Large     &stage1                    &61.43            \\
   Decoder A                           &BEiT-Large     &stage2                    &\textbf{61.44}            \\
   Decoder A                           &BEiT-Large     &stage3                    &61.43   \\
   \hline
   \hline
\end{tabular}}
\end{table}

\section{Experiments and Analysis} \label{exp}

\subsection{Training Configuration}
Our method is implemented in PyTorch ($version \geq 1.3$) \cite{paszke2019pytorch} and trained on 8 NVIDIA Tesla V100 GPUs with a 32 GB memory per-card.
The overall learning consists of two stages: pre-training stage and fine-tuning stage.

\noindent \textbf{pre-training stage.} We learn the backbone network of our framework by leveraging ImageNet22K dataset \cite{deng2009imagenet} following the settings described in \cite{bao2021beit}.

\noindent \textbf{fine-tuning stage.} The initial learning rate is set as $0.00002$ and the weight decay is $0.05$. 
We set the crop size of the input image as $512 \times 512$ and batch size as $16$ by default.
Besides, the networks are fine-tuned for $240$ epochs on the train set. For each iteration, we randomly select one frame from the videos to train our framework.

\subsection{Abaltion Study}
\noindent \textbf{Model Ensemble.} As indicated in Table \ref{table1}, the designed Decoder A outperforms the baseline model by $1.22\%$ mIoU and the proposed Decoder B is $1.16\%$ mIoU higher than the baseline model.
Finally, by the technology of model ensemble, we obtain the final video segmentation framework with a mIoU of $62.12\%$.

\noindent \textbf{Multi-stage Inference.}
As illustrated in Table \ref{table2}, we show the ablation study on the multi-stage inference. To be more specific, the stage0 means the original output $Mask_{0}$ of Decoder A. And stage1 means the predicted segmentation mask $Mask_{1}$ by replacing $\mathcal{P}$ with $Mask_{0}$.
Similarly, stage2 and stage3 use $Mask_{1}$ and $Mask_{2}$ to replace the $\mathcal{P}$ to obtain the predicted segmentation mask, respectively.

\subsection{Video Segmentation Framework}
Our final video segmentation framework is built by ensembling the Decoder A and Decoder B as well as leveraging the trick of multi-stage inference and we finally achieve a mIoU of $57.44\%$ on the test set of VSPW.

{\small
\bibliographystyle{ieee_fullname}
\bibliography{egbib}
}

\end{document}